# Machine Learning in Network Centrality Measures: Tutorial and Outlook[1]

FELIPE GRANDO, Institute of Informatics, Federal University of Rio Grande do Sul, Porto Alegre, Brazil, fgrando@inf.ufrgs.br
LISANDRO Z. GRANVILLE, Institute of Informatics, Federal University of Rio Grande do Sul, Porto Alegre, Brazil, granville@inf.ufrgs.br
LUIS C. LAMB, Institute of Informatics, Federal University of Rio Grande do Sul, Porto Alegre, Brazil, lamb@inf.ufrgs.br

Complex networks are ubiquitous to several Computer Science domains. Centrality measures are an important analysis mechanism to uncover vital elements of complex networks. However, these metrics have high computational costs and requirements that hinder their applications in large real-world networks. In this tutorial, we explain how the use of neural network learning algorithms can render the application of the metrics in complex networks of arbitrary size. Moreover, the tutorial describes how to identify the best configuration for neural network training and learning such for tasks, besides presenting an easy way to generate and acquire training data. We do so by means of a general methodology, using complex network models adaptable to any application. We show that a regression model generated by the neural network successfully approximates the metric values and therefore are a robust, effective alternative in real-world applications. The methodology and proposed machine learning model use only a fraction of time with respect to other approximation algorithms, which is crucial in complex network applications.

## 1 INTRODUCTION

Complex network models are ubiquitous to several computer science domains (*e.g.*, [1] [2] [3] [4]). This motivates the development and application of several metrics for their understanding, analysis, and improvement. Moreover, the computational analysis of such networks in real-world environments is a fundamental tool in many fields such as mobile and 5G wireless networks [5] [6].

Some of the most widely used network measurements aim at the evaluation, ranking, and identification of important vertices by their power, influence, or relevance. They are usually known as vertex centrality measures [7] [8] [9]. These metrics constitute a class of metrics and underlying algorithms, capturing a different idea of centrality, which is used in typical and fundamental applications.

Even though the metrics' algorithms are polynomial in time, the computation of such metrics becomes a difficult, often complex problem when they are applied to real-world networks, composed of thousands or even millions of elements and their connections. The use of centrality measures in large-scale networks and real-time applications demands principled approaches. Moreover, many application environments deal with dynamical networks that are constantly changing. Such applications make use of snapshots of their networks to analyze properties and changes over the time, which requires the computation from scratch of the network analysis metrics, such as centrality measures, for each network snapshot.

In this tutorial, we show how to use and apply a methodology and associated machine learning-based techniques to effectively approximate vertex centrality measures. We do so by using neural networks learning in order to build a regression model. To guide and show the effectiveness of the methods, we apply the techniques to real world networks using two of the most important vertex centrality measures: *betweenness and closeness* centralities.

We show how to use fast and feasible training methods for artificial neural networks, where training data are obtained with a complex network model called Block Two-Level Erdős and Rényi – BTER [10] [11]. The BTER model generates networks with diminished size, but with the same properties of the huge analyzed networks. This allows both researchers and practitioners unlimited an easy access to training data for whatever application one may have to tackle. Such a methodology is capable of providing enough information for the neural network to be able to generalize for the real datasets comprising huge networks.

We illustrate our tutorial with several configurations for the neural networks, including many standard machine learning algorithms, different network structures, and combinations of meta-parameters. In this way, we identify the appropriate setup for such tasks. The tutorial shows how to use the configuration that presents the best quality and performance results in a set of 30 (large scale) real-world networks. Finally, we compare the results obtained from the machine learning model with the exact computation of the centrality measures and other approximation techniques proposed in the literature. We show how the machine learning methodology produces competitive results with quality comparable with other approximations methods but – perhaps more importantly - at just a fraction of time and space. The methodology and model used in this tutorial are capable of computing the ranking assumed by the centrality measures in linear time/space requirements and under reduced error margins.

The remainder of this tutorial paper is organized as follows. First, we present key applications of centrality measures and then a prolegomenon to vertex centrality measures and complex network models, highlighting key definitions. Next, we present approximation methodologies for network centrality measures, including a systematic tutorial on how to train, test, and use machine learning for such purposes. We then evaluate and compare results, by presenting several applications of centrality measures that can benefit from the machine learning based methodology. Finally, we conclude and outline directions for further research in the field.

---

[1] Preprint accepted at **ACM Computing Surveys,** vol 51(5) article 102, 2018. https://doi.org/10.1145/3237192 This research was financed in part by CAPES Fundação de Coordenação de Aperfeiçoamento de Pessoal de Nível Superior, Finance code 001 and by CNPq: Brazilian National Research Council.

## 2 BACKGROUND AND RELATED WORK

In this section, we briefly outline applications where centrality measures play a key role. In particular, we motivate the use of machine learning and approximation techniques that shall be useful when tackling large-scale complex networks.

### 2.1 An Outlook of Centrality Measures

Centrality measures are used as important analyses tools in several applications [12] [13] [14]. They are usually employed as a mean to identify relevant and important elements by identifying their behavior and roles within given complex networks. Centrality measures are also effective as a comparison factor with other domain-specific metrics and can be used as a bridge to other areas [15] [16] [17] [18].

In the coming subsections, we highlight promising areas in which recent studies have used centrality measures as a meaningful instrument. We do so by a simple taxonomy, classifying such examples considering their main application areas – of course, some of the researches are related to more than one application domain.

### 2.2 Centrality in Computer Networks

Given the ubiquity of computer networks structures, the use of centrality measures is pertinent as heuristics to improve the solution of security problems, control issues, communication flow, and resources optimization.

One of many examples is the study of Maccari and Lo Cigno [19], who applied the betweenness centrality in the optimization of online routing control protocols responsible to provide a fast and efficient recovery from a node failure with minimal disruption of routes. They consider that the failure of a node with high centrality (they have focused in the top 10 ranked vertices) generates higher loss compared to the failure of peripheral nodes and applied this fact in their proposed optimization problem formula to minimize routes' disruption with reduced control message overhead. They also pointed it out that, although the centrality needs to be computed online (it takes about seconds to compute) and this is feasible only in networks with hundreds of vertices, there are approximation techniques that may be used for networks with thousands or more vertices. Moreover, the same technique can be extended to distance-vector protocols that are largely used in wireless mesh networks [20]. Ding and Lu [21] also studied nodes' importance (given by centrality measures) to maintain the structural controllability of a network.

Another key application of centrality measures is in wireless community networks (WCNs). WCNs are bottom-up broadband networks empowering people with their on-line communication means [22]. However, WCNs often lack in performance because services characteristics are not specifically tailored for such networks. Centrality measures are used as heuristics that improve the selection of peers and chunks within WCNs in the communication between users to reduce the impact of P2P streaming while maintaining applications performance.

Kas *et al.* [23] studied social network analysis metrics and centrality measures as tools to improve the design of wireless mesh networks with better performance by efficiently allocating the resources available. The allocation of network resources using the centrality measures as heuristics has been a topic of study in distributed placement of autonomic Internet services for the efficient support of users' demands [24].

There are also studies about mobile and 5G wireless networks that used centrality measures to reduce network traffic [5] [6]. Likewise, centrality measures are studied in the context of network security. For instance, they have been used to configure distributed firewalls [25] and to identify critical nodes in which intrusion detection and firewalling is most suitable [26]. In addition, they were applied to build cloud-based filters and footprints of traffic inspection algorithms at global scrubbing centers [27].

### 2.3 Complex Networks Construction

The analysis and understanding of the structure of complex networks are fundamental to understand variables and properties that contribute to the network formation and to identify which premises affect network development and evolution [1] [3]. Centrality measures help these studies by identifying nuances and characteristics of components of the networks. In such context, centralities are used as means of understanding the role of a given element and its effect on the other elements of the entire network.

König *et al.* [28] analyzed the underlying differences between technological and social networks and the assortativity behavior of network elements. They considered that the link formation is mainly based on the vertex centrality. Centrality measures were also used as part of a dynamic network formation model to explain the observed *nestedness* in real-world networks [29] and social networks community structures [30].

They are also fundamental to several community detection and formation algorithms [31]. The betweenness centrality measure was applied successfully by Newman and Girvan [31] as a heuristic metric in order to define proper cuts/divisions in social networks as a way to separate the elements and iteratively search for communities. The idea behind the algorithm proposed by the authors is that the more central elements are bridges between different communities. As an iterative algorithm, it needs to compute the centrality measure again at each step/cut/division made, and because of its frequent use in huge social networks an efficient implementation is required. Therefore, approximation methodologies are preferred most of the time.

**2.4 Artificial Intelligence Applications**

The use of network models in communication, cooperation, or learning structures is key in AI. Several AI domains offer themselves to the use of centrality measures as means of strategy, guidance, or simply as domain information. Machine learning algorithms and methods can use centrality measures as heuristics to learn faster, to synthesize input data, or as additional information about an AI application, which in turn helps in the generalization for a given domain.

Pedestrian detection is one of the example problems that used centrality measures associated with machine learning techniques. Detecting pedestrians in computer vision is a critical problem when one considers that the performance of trained detectors may drop very quickly when scenes vary significantly, which is frequently the case. Cao *et al.* [32] proposed a novel methodology for such a problem based on a bag of visual words to detect pedestrians in unseen scenes by dynamically updating the keywords using centrality measures to select new keywords on the manifold model.

Another AI application used the metrics to classify label-dependent nodes in a network (e.g. hyperlinks connecting web pages), citations of scientific papers, e-mail conversations, and social interactions in the Web [33].

Centrality measures can be applied in several ways as an optimization heuristic [34] in multi-agent systems and multi-robot teams, since they are essentially driven by their communication organization and strategies. Xu et al. [34] noticed through simulations that the more central robots (considering its betweenness centrality) in a large robot team (organized frequently as clusters) are the ones responsible to broadcast the information amongst clusters. Consequently, these central elements play an important role in group efficiency and are helpful to speed up the information diffusion.

In addition, reinforcement learning applied to agents' automatic skill acquisition has also been studied using centrality measures [35] [36].

**2.5 Social Network Analysis**

There is a growing amount social networks data, from different sources. Therefore, it is crucial to study the available information, as many networks comprise hundreds of millions of vertices. Centrality measures are one of the most important analyses tools for these kinds of networks and can be used for many purposes and fundamental comparisons, which offer insights on their social impact and behavior.

Danowski and Cepela [37] used centrality metrics to analyze the impact that presidential centrality role has on presidential job approval ratings. They hypothesized that when the centrality of the president is lower than of the other cabinet members, job approval ratings is higher. Jiang *et al.* [38] introduced a model to find influential agent groups and their impact in multiagent software systems by using centrality measures, while Mcauley and Leskovec [39] used the metrics to search and identify social circles in ego networks.

The identification of important elements and an impact analysis of coautorship networks [40] is also a research topic where centrality measures proved to be fundamental. They have used data from 16 journals in the field of library and information science with a time span of twenty years to construct a coautorship network and, they tested the use of four centrality measures (closeness, betweenness, degree and PageRank) as predictors of articles' impact. To validate such an analysis, they compared the results obtained with the centralities with the citation counts. They unveiled a high positive correlation amongst it and the metrics, which strongly suggests that central authors are mostly likely to have a better reputation and produce articles with increased citations. Louati *et al.* [41] applied the metrics as selection heuristics to desired trustworthy services for social networks. Likewise, there are studies about the setup of marketing strategies involving the selection of influential agents that increase the impact factor and improve the match between marketing content and users' expectation and interests have also proved the need for centrality measures [42] [43].

**2.6 Traffic and Transport Flow**

Physical transport of goods and people are a huge strategic logistic problem for cities, states, and countries. For instance, the analysis of terrestrial, air, and water traffic networks used centrality measures as heuristics to solve flow and routing problems. Hua [44] analyzed the United States air transportation network to unveil the importance and impact that each airport has over the air traffic network as a whole.

Gao *et al.* analyzed the potential of centrality to predict road traffic and flow patterns [45]. They examined urban street flow using the taxis trajectory (recorded via GPS) and compared to the result predicted using betweenness centrality and the consideration of spatial heterogeneity of human activities, which was estimated using mobile phone Erlang values. The combination of centrality with other techniques showed to be extremely effective.

Centrality measures and their models are also used in global and local scenarios [46] and on the analysis of different points of view (intersection, road, and community views) in such scenarios [46].

**2.7 Centrality in Game Theory**

In game theory, centrality measures are employed in studies about coalitional games. Coalitional or cooperative games are usually formed by groups of individuals that are connected via a network. The efficiency of this network is often related to the groups' performance [47].

Noble et al. [47] studied the impact and relationship between the centrality of an agent with their collective performance in the resolution of cooperative problems. They tested several centrality measures (betweenness, closeness, degree, and eigenvector) combined with different network structures (defining communication channels among agents) and distinct solving strategies (evolutionary algorithms) in a set of minimization problems (real-valued functions). They showed that the centrality of an agent severely affects its contribution for the overall collective performance of the network.

Centrality measures also have shown significance to define strategies for the dynamic formation and evolution of networks connecting individuals in social and economic game-based networks. Further, the measures have also been applied to risk assessment, considering fairness and efficiency, and to identify a bank's systemic impact and relevance [49]. Most games also make use of complex centrality measures, and as they require efficient algorithms and sometimes approximation algorithms to be feasible [48].

### 2.8 Biological Networks

Biological networks are examples of network structures not built by humans composed by biological structures. Protein-protein interaction (PPI) is one particular type of biological network where centrality measures have been applied [50]. They have applied centrality measures like betweenness, degree and closeness to select proper/informative candidates (proteins) for labelling in PPI networks clusters and then classify/predict their purpose or function inside a biological cell. They reason that the more central proteins in a PPI network cluster are probably responsible for a given cell function. The identification and the behavior of important proteins related to specific reactions in a cell constitute a huge step towards the development of drugs with minimal collateral effects and also to better understand diseases and hormonal interactions and their effects. Note that PPI networks are usually huge, sparse, and numerous. So, an efficient analysis method is fundamental in these applications.

The metrics were also important in several studies about human brain connectivity and its behavioral interpretations [51] [52]. Neural connections and interactions (synapses) form huge complex networks where actions of a living being is controlled and determined. Therefore, the identification, classification, clustering, behavior, and inner-relationship are topics of high research interest. There is a growing demand for the applications of centrality measures in the biomedical sciences due to their successful use in many network related topics and the insights obtained from several application domains [51] [52].

## 3 PROLEGOMENON TO CENTRALITY MEASURES AND COMPLEX NETWORKS

In this section, we cover the main concepts underlying the most important centrality measures and complex network models that shall be used in the sequel. Both topics are vital to understand of applications involving social, biological, and technological complex networks.

Centrality measures are fundamentally complex and diverse, and so is the field of complex networks. Therefore, we provide brief, but clear explanations about their underlying definitions, ideas, purposes, and algorithms. For more comprehensive surveys on centralities, please see e.g. [53] [54]; complex networks and graph measurements are surveyed in, e.g. [55] [56].

### 3.1 Vertex Centrality Measures

Vertex centrality measures can be characterized by deterministic algorithms (and associated formulations) that describe the computation of a "centrality" value for each vertex of a graph or network. However, the concept of what centrality is and what exactly it measures depends on its mathematical formulation. As a result, there are several centrality measures capturing distinct ideas of centrality. Many of them are very similar to each other in their conception and computed results, while others are specifically designed to networks in a determined field [7] [8] [9] [13].

Informally, the main question that the centrality measures try and answer is how central a vertex is, how much it is at the core or in the periphery of a network. Depending on the definition of centrality, an individual with high centrality is the one that has more control, visibility, or independence inside a network. There are metrics suitable to directed and weighted graphs, while others are used in strong connected simple graphs, or in temporal graphs. Vertex centrality measures make use only of the structural pattern of the network (*i.e.*, vertices connections, edges weights, and temporal timeline) for their evaluation. Therefore, all formulas and algorithms receive as input an adjacency matrix or adjacency list, with weights and temporal sequence when appropriate, representing the network under analysis.

For certain centrality measures, meta-parameters can be tuned to adapt the metric to a given kind of network structure or to control the weights given to different aspects/parts of the centrality. The metrics adapted to directed graphs are examples of the former case and the ones composed of more than one centrality measure are examples of the latter case.

In this section, we detail four of the main vertex centrality measures: *betweenness, closeness, degree, and eigenvector*. We also briefly explain some of the most relevant metrics in the literature. In the tutorial, we focus on unweighted symmetric networks, so we present the metric's versions adapted for such kind of graphs; however, some modifications turn some of the metrics suitable for most kinds of graphs. Notice that in most applications where centrality is important, the vertices rank is used instead of their absolute centrality value. This aims at better comprehension, interpretation, and comparison of values. Rank centrality is a post-processing of the centrality measures where, for each vertex, a rank is given based on their centrality value decreasing order. If vertices are tied, their resulting rank is averaged.

Centrality measures can be classified into four groups with respect to their purpose and conceptualization.

*3.1.1 Degree Centralities*

These are the simplest and most straightforward centrality measures. They were introduced by Shaw [7], formally defined by Nieminen [8] and popularized by Freeman [9]. These centralities are related with the idea of visibility that a vertex has among its neighbors.

The degree centrality ($C_D$) of a vertex $w$ is the number of edges connected to it, *i.e.*, the number of its adjacencies (1) [57].

$$C_D(w) = \sum_{i=1}^{n} a(i,w) \qquad (1)$$

The corresponding algorithm has time complexity $\Theta(m)$ by counting each edge and does not require any specific graph representation [58]. So, an adjacency list can be used for sparse graphs with space complexity $\Theta(n+m)$ or an adjacency matrix for denser graphs with space complexity $\Theta(n^2)$ [57]. The degree centrality is the less complex and has the smallest computational cost among centrality measures.

The metric is subdivided into indegree and outdegree for directed graphs. There are variations in which the weights of the edges are accounted for. This class of metrics is the only one that considers only local information to evaluate each vertex centrality. Notwithstanding, it is highly correlated to all other centrality measures despite its simpler formulation. This metric has low capability at distinguishing vertices or ranking them, *i.e.*, it considers many vertices as equally central [12].

*3.1.2 Path Centralities*

This group of centralities evaluates the vertices as being central if they are in between (or at the "crossroads") of many "paths". This fact allows the vertices to control the communication through such paths. Each centrality of this class considers different kinds of paths or consists of a distinct evaluation of these paths.

Most of these metrics require the graph to be strongly connected or evaluate each connected component of the graph individually and independently. However, there are more tolerant variations or adaptations that relax these restrictions to any kind of graph structure.

The most widespread and used metric is the *betweenness* centrality ($C_B$), which considers only the smallest paths, called geodesics. This concept was introduced by Shaw [7] and formally defined by Freeman [13].

The betweenness of a vertex $w$ is the number of geodesics between vertices $i$ and $j$ that passes through vertex $w$ divided by the total number of geodesics between the vertices $i$ and $j$ (2) [57]. Considering $i$ and $j$ all vertices of the graph, and $j$ larger than $i$ (a geodesic from $i$ to $j$ and from $j$ to $i$ is considered only once) [57].

The metric can be calculated using Brandes' algorithm [14], keeping its time complexity at O(mn) and does not need any specific representation for the input graph [57] [58]. It is the least complex metric in this class of centrality measures.

$$C_B(w) = \sum_{i=1}^{n} \sum_{j=i+1}^{n} \frac{g_{ij}(w)}{g_{ij}} \qquad (2)$$

Other examples of centrality measures that belong this class of metrics are:
 (i) Flow betweenness [15]: based on the network's maximum flow paths;
 (ii) Random walk betweenness [16]: considers all paths, but weighs them according to the probability in which a random walk would passes through it;
 (iii) Communicability betweenness [17]: considers all paths, but rates the longest paths as less important;
 (iv) Range limited betweenness [18]: considers only the shortest paths inside a range limit.

*3.1.3 Proximity Centralities*

The basic idea of proximity centralities is that the lower the distance between a vertex to the others, the higher its centrality value and its independence from the network. The main difference among these centralities is that each metric computes the "distance" between vertices in a distinct way. Since these centralities are based on distance metrics, there is an inherent problem with disconnected graphs: depending on the centrality measure, the distance between two disconnected vertices are considered infinite or the largest possible distance for the given network size.

The most prevalent and used centrality measure of this group is the *closeness centrality* ($C_C$), first presented by Bavelas [59] and then formally defined by Sabidussi [60]. Closeness centrality is the sum of the geodesics inverse distances from the vertex analyzed to all other vertices (3) [57] [58].

$$C_C(w) = \frac{1}{\sum_{i=1}^{n} d(i,w)} \qquad (3)$$

This metric can be calculated using a small variation of Brandes' algorithm [61], keeping the same time complexity of betweenness centrality – and it does not require any specific representation for a graph [57] [58].

Other examples of centrality measures that belong to this class are:
 (i) Informational centrality [62]: it is computed by the probability that a random walk starting from the start point ends in the target point;
 (ii) Eccentricity [63]: considers only the distance to the farther vertex from the starting point;
 (iii) Random walk closeness [64]: the distance is measured by the average random walk time it takes to arrive to a target;

(iv) Hierarchical closeness [65]: it is computed by a composition of closeness centrality and degree centrality (out degree in directed graphs).

*3.1.4 Spectral Centralities*

All metrics that belong to this group evaluate the vertices centrality by their participation in substructures of the network. They are called spectral measures because of their relation with the set of eigenvalues of the adjacency or Laplacian matrix of the graph representing the network. Perhaps the mostly widely used among these measures is the *eigenvector centrality* ($C_E$). Bonacich [66] suggested the centrality based on the eigenvector of the largest eigenvalue of a network's adjacency matrix. Eigenvectors can be seen as a weighted sum of not only immediate contacts but, as well as, indirect connections with every vertex of the network of every length [57] [58] [66]. Moreover, it weighs contacts of a vertex according to their own centrality, *i.e.*, links with central vertices contribute towards their own centrality.

The eigenvector respective to the largest eigenvalue can be computed via an iterative procedure known as "power method" using the adjacency matrix and an auxiliary vector [67], which reduces its computational cost considerably and avoids numeric precision issues [57].

The power method requires an infinite number of steps (worst case) but as the number of steps increases, the precision of the measure also increases [57]. Therefore, the number of decimal places can be used to turn this measure feasible even for massive networks where a hundred steps usually grants enough precision to differentiate the vertices centrality values [57] [67].

The eigenvector centrality value of a vertex $w$ at an iteration $it$ is the $w$ index of a vector $E$ multiplied by the adjacency matrix $A$ divided by the sum of the elements of $E$ at a previous iteration (4) [57] [67]. The vector $E$ can be initialized with any real positive number [57] [67].

$$C_E(w) = \sum_{it=1}^{+\infty} \frac{(E^{it}A)_w}{\sum_{i=1}^{n} E_i^{it-1}} \quad (4)$$

The following centrality measures also belong to this class of metrics:
(i) Katz centrality [68]: similar to the eigenvector centrality, but uses an attenuation factor to reduce the influence of distant vertices;
(ii) PageRank centrality [69]: similar to the Katz centrality, but adds a dumping factor applied to directed graphs (correcting the rank sink and drain problems caused, respectively, by cycles and vertices with zero outdegree);
(iii) Subgraph centrality [70]: evaluates the vertices by their participation in subgraph structures of the network giving less weight as the subgraph becomes larger;
(iv) Functional centrality [71]: similar to the subgraph centrality, but with limited range on graph structures.

In Table I we present a summary of the characteristics of the centrality measures discussed in this subsection. It shows if the centrality measures can be applied to digraphs and/or graphs with weighted edges; the expected parallelism speedup due to algorithmic restrictions and dependencies; the granularity of each measure, *i.e.*, its ability to distinctly evaluate the vertices; and the complexity of the centrality algorithm and whether its parameters have to be tuned manually.

It is important to notice that although Eigenvector, Katz and PageRank exact algorithms have quadratic upper bounds considering their time complexity, they perform in practice near linear time due to the application of an iterative version of the algorithms ("power method" [67]). For instance, this fact enabled the use of such centralities in the core of the ranking system of Web search engines, such as Google [69].

TABLE I. CENTRALITY MEASURES SUMMARY

| Centrality Measure | Digraph | Weighted Graph | Parallelism Speedup | Granularity | Parameterized | Complexity |
|---|---|---|---|---|---|---|
| Degree | No | Yes | High | Low | No | $\Theta(m)$ |
| Indegree | Yes | Yes | High | Low | No | $\Theta(m)$ |
| Outdegree | Yes | Yes | High | Low | No | $\Theta(m)$ |
| Betweenness | No* | No* | High | High | No | $O(mn)$ |
| Flow Betweenness | No* | Yes | Medium | Low | No | $O(m^2n)$ |
| Random Walk Betweenness | No | No | Low | High | No | $O(mn^2+n^3)$ |
| Communicability | No | No | Low | High | No | $O(n^4)$ |
| Range Limited Betweenness | Yes | Yes | Low | High | Yes | $O(mn^{1+l/d})$ |
| Closeness | No* | No* | High | Medium | No | $O(mn)$ |
| Information | No | Yes | Low | High | No | $O(n^3)$ |
| Eccentricity | No* | No* | High | Low | No | $O(mn)$ |
| Random Walk Closeness | No | No | Medium | High | Yes | $O(n^2)$ |
| Hierarchical Closeness | Yes | No* | High | Medium | No | $O(mn)$ |
| Eigenvector | No | No | Medium | High | Yes | $O(n^2)$ |

| Centrality Measure | Digraph | Weighted Graph | Parallelism Speedup | Granularity | Parameterized | Complexity |
|---|---|---|---|---|---|---|
| Katz | Yes | No | Medium | High | Yes | $O(n^2)$ |
| PageRank | Yes | Yes | Medium | High | Yes | $O(n^2)$ |
| Subgraph | No | No | Medium | High | No | $O(n^2)$ |
| Functional | No | No | Medium | High | Yes | $O(n^2)$ |

*. There is a more complex version of the algorithm that supports this characteristic.

### 3.2 Complex Network Models

Recently, complex network research has aimed not only to the identification and understanding of network principles, but also to the effective use of their properties and applications [1] [2].

Research results have shown that technological, biological, and social networks share common properties, such as low diameter, high clustering coefficients, presence of community structure, and scale-free degree distribution [64] [65]. These developments have led to the identification of complex network models capable of stochastically generate networks with similar or related properties. These models were proposed with two main goals: (*i*) to understand what underlying effects give rise to such properties, and (*ii*) to produce synthetic networks with controlled characteristics that may serve as research tools for many disciplines [3] [12] [72] [73].

In this subsection, we briefly summarize the most well-known complex network models, starting by the simpler and elder ones.

#### 3.2.1 Erdős and Rényi Random Graphs

Erdős and Rényi [74] introduced the first model that generates simple random graphs [58]. It defined a fixed number of vertices $n$ and a probability $p$ of connecting each pair of vertices, which also corresponds to the final clustering coefficient of the graph [58]. The higher is the value of $p$, the higher the mean degree and density, and the lower the diameter of the network [58]. These simple graphs do not represent real-world networks with high fidelity because they do not present any community structure and because their degree distribution follows a Poisson distribution [58].

#### 3.2.2 Watts and Strogatz Small-World Model

In small-world networks, most vertices can be reached within short paths [58]. In addition, these networks show a large number of small cycles, especially of size three [58]. Watts and Strogatz [72] proposed a model to generate networks with the small world properties. The graph starts with a ring of connected vertices, each one adjacent to its $k$ nearest neighbors [58]. Then, with probability $p$, each edge is randomly reassigned to any available position. This relinking method, with an intermediate or small $p$ (typically $p$ should be lower than 0.5), will create paths among distant vertices while keeping a high clustering coefficient among close neighbors [58].

The higher is the value of $k$, the higher the vertex mean degree, clustering, and density, although diameter decreases. In addition, the higher is $p$, the lower is the clustering coefficient and diameter [58].

#### 3.2.3 Barabási and Albert Scale-free Networks

The analyses of large social networks data show that their degree follows a scale-free power-law distribution. Barabási and Albert [73] explained this property using the fact that networks expand continuously by the addition of new vertices and that these new vertices attach preferentially to vertices already well connected, *i.e.*, vertices with higher degree (or "the rich get richer") [58].

The model proposed by them has the above features. It starts with $k$ number of fully connected vertices and keeps adding new vertices with $k$ connections, defined by a preferential attachment formula [58]. The probability of a vertex $p_i$ receiving a new connection takes into consideration the degree $d$ of the vertex divided by the sum over the degree of all vertices. In this way, high degree vertices have a greater chance of receiving new connections than vertices with lower degree: the higher values of $k$, the mean degree, clustering coefficient, and density gets higher, while the diameter sinks [58].

#### 3.2.4 Networks with Community Structure

Newman and Park [30] analysis of several social networks showed that such networks are formed by community structures. Each vertex has many connections with vertices inside the same community and few connections with vertices of other, *i.e.*, outside communities [58]. In addition, they discovered that, in such networks, high degree vertices tend to be connected with other high degree vertices too [58]. Further, they showed that vertices with small degree are usually connected with vertices with small degree (*i.e.*, they present dissortativity behavior) [58].

They proposed a model that generates random networks with the above properties [58]. Their model starts defining $c$ communities and an (uneven) distribution of vertices for each community that represents distinct group sizes [58]. Further, each vertex can be assigned to more than one community [58]. Then, each vertex has a fixed high probability $p$ of being connected to every other element of its own communities and zero probability to be connected with vertices to which it does not share a community [58]. Notice that the vertices that were assigned to more than one community are those that link the different communities [58]. In this network model, the higher the value of $p$ and the lower the value of $c$, the higher is the network's mean degree, clustering, and density, although network diameter decreases [58].

*3.2.5 Geographical Models*

Complex networks are generally considered as lying in an abstract space, where the position of vertices has no definite particular meaning [58] [55]. However, several kinds of networks model physical interactions in which the positions of the vertices characterize a higher probability of interaction with close neighbors than with distant ones [58]. Costa *et al.* [55] introduced a model of this behavior where the probability of vertices $i$ and $j$ being connected decays exponentially with distance between $i$ and $j$ [58].

*3.2.6 Kronecker Graphs*

Leskovec *et al.* [75] proposed a model in which real world networks are formed by the same substructures, repeatedly and recursively. The model defines an algorithm that estimates parameters capable of generating synthetic networks with properties similar to any given real network.

The model starts with an adjacency matrix typically of order two to four. Each cell of the matrix represents the probability that a vertex has a link towards the other in a core substructure. An algorithm proposed by the same authors in which a real network is used as basis, estimates these initial parameters. The matrix is symmetrical for undirected graphs and asymmetrical for directed graphs. To recreate networks of arbitrary size, the matrix is multiplied by itself using the Kronecker product, a generalization of the outer product. The Kronecker product doubles the size of the network and can be repeated until the desired size is achieved. The resulting matrix is composed of several parameters (probabilities) that can be used to generate networks with properties that approximate the real network given as basis.

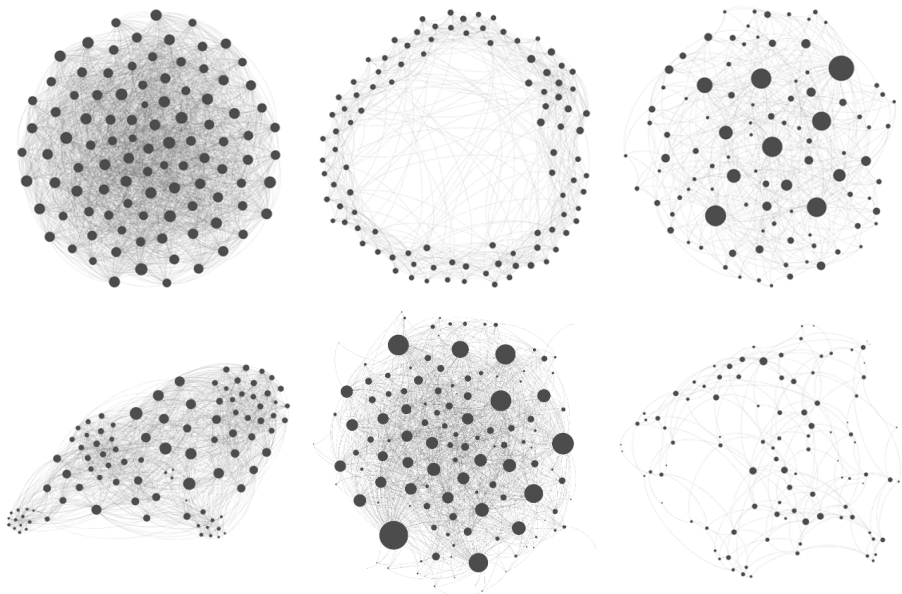

Fig. 1. Sample networks generated by the complex network models (simple random graph, small-world model, scale-free model, networks with community structure, geographic model and Kronecker graphs, respectively, from top-left to bottom-right). The size of the vertices is proportional to their degree. All networks contain about a hundred vertices. Source [58]

*3.2.7 Block Two-Level Erdős and Rényi Model*

The Block Two-Level Erdős and Rényi (BTER) model [10] generates networks with very similar properties of real networks [57]. It builds a network based on a degree distribution and a desired clustering coefficient. All networks generated by the BTER model present community structures and low diameter (*i.e.*, small-world phenomena) [10] [57].

The BTER model is divided into three steps [11]. First, the vertices are grouped by degree in communities with size equal to the degree of its members plus one [57]. Then, each community is considered an Erdős and Rényi graph where the probability of connection among vertices is equal to the desired clustering coefficient [57]. The last step generates connections proportional to the excess degree of each vertex (number of connections that a vertex needs to match its desired degree) [57]. This weighted distribution is based on Chung and Lu graphs [76] [77].

TABLE II. COMPLEX NETWORK MODELS SUMMARY

| Network Model | Degree Distribution | Assortativity* | Number of Parameters | Configuration Complexity |
|---|---|---|---|---|
| Erdős and Rényi | Poisson | Positive | 2 | Low |
| Small-World Model of Watts and Strogatz | Dirac Delta | Positive | 3 | Low |
| Scale-free Networks | Power Law | Positive | 3 | Low |

| Network Model | Degree Distribution | Assortativity* | Number of Parameters | Configuration Complexity |
|---|---|---|---|---|
| Networks with Community Structure | Multinomial | Negative | At least 3 | Medium |
| Geographical Models | Poisson | Negative | At least 2 | Low |
| Kronecker Graphs | Lognormal | Any | At least 5 | High |
| Block Two-Level Erdős and Rényi | Multinomial | Any | At least 3 | Medium |

*. The assortativity coefficient or assortative mixing is the Pearson correlation coefficient of the degree between pairs of connected vertices. A positive assortativity means that vertices tend to be connected with vertices with similar degree while a negative coefficient means that vertices are connected with higher or lower degree vertices.

In Table II we present a summary of the characteristics of the complex network models discussed above. It shows the expected degree distribution and the assortativity coefficient of vertices degree in the networks generated with the model. It also presents the number of configurable parameters of each model and the overall complexity to configure such parameters to generate networks with desired characteristics (considering the restrictions and capabilities of each model).

## 4 MACHINE LEARNING METHODS FOR NETWORK CENTRALITY MEASURES

Several authors have proposed approximation techniques for specific network metrics. Unfortunately, typical centrality measures algorithms do not scale up to graphs with billions of edges (such as large social networks and Web graphs). Even though their algorithms are polynomial in time, their computation requires days or even months for massive networks [78] [79]. Closeness metric, for instance, takes about 5 years to compute for a network with 24 million vertices and 29 million edges [80]. This can be even more critical when one needs to compute many centrality queries, particularly when one is interested in the centrality of all vertices or whenever the network structure is dynamic through the time. This feature is common to most real networks, which are constantly changing and evolving in time [1] [55].

In this section, we describe the most common method for sampling and then a detailed tutorial of the methodology based on machine learning techniques for network centrality learning and computation.

### 4.1 Vertices Sampling Techniques

The underlying premises behind a sampling technique is simple: one should compute the exact centrality value for a predefined number of sampled vertices and then estimate the centrality value of the others based on such computation. For instance, the betweenness and closeness centralities share a quite similar foundation. One computes the single source shortest path (SSSP) for a given number of sample vertices. Each SSSP tree gives the exact centrality value for its source vertex. At the same time, one can use them as an approximation for the other vertices considering that all previously computed SSSP trees are partial solutions for such vertices. Therefore, a given vertex not sampled will have its centrality value approximated by an average result given by all SSSP trees from the sampled vertices. An algorithm for such objectives was defined and tested in real case scenarios by Bader *et al.* [78] for betweenness and by Eppstein and Wang [79] for closeness centralities.

However, the simple approach given by the sampling technique has a few drawbacks and leads to relevant questioning such as how many vertices should be sampled and how should one select them? Or, how can one efficiently parallelize the sampling technique considering that it is not possible any longer to compute each vertex centrality independently?

Brandes and Pich [81] studied the vertices selection problem. They proposed several heuristics to choose vertices for the sampling techniques, starting with simple ones, such as picking the high degree vertices and finishing with more complex ones, which considers vertices distances and mixed strategies. Despite all their attempts to find an optimal heuristic for such problem, they concluded that picking vertices uniformly at random on average is the best strategy when considering different types of network structures.

### 4.2 Building Sampling Experiments

In order to make the concepts clearer to the reader, we put the ideas into a practical experimentation. This will serve to illustrate the capabilities of the method in validation and comparative analyses.

We shall illustrate the effect of sample sizes in a parallelized version of the sampling algorithms in real case scenarios using 30 real-world networks from four freely available data repositories. The selected networks are symmetric and binary (unweighted edges). Only the largest connected component (LCC) was used for the tutorial experiments, which was computed for all analyzed networks. The list of all real networks with their respective size and the size of its LCC (proportion w.r.t. the number of vertices), classified by network type and grouped by data repository is presented in Table III.

In the sequel, we computed each of the four exact centrality measures (eigenvector – $C_E$, betweenness – $C_B$, closeness – $C_C$, and degree – $C_D$) for all real networks. The computation of eigenvector and degree centralities is sequential, while betweenness and closeness centralities are computed simultaneously with a merged algorithm keeping the same time complexity upper bounds and using parallelism [61]. All algorithms[2] were programmed in C and the parallel computations used the native OpenMP (Open Multi-Processing interface).

The computation of the metrics used a SGI Altix Blade with 2 AMD Opteron dodeca-core with 2.3GHz, 128KB L1 Cache and 512KB L2 Cache per core, 12MB L3 cache total and 64GB DDR3 1333MHz RAM memory, Red Hat Enterprise Linux Server 5.4.

---
[2] The source code of the experiments in this tutorial paper is freely available at https://github.com/fgrandoinf/centrality-measures

The computation time is presented in Table IV, and represents the actual time spent for each of the metrics. Degree centrality required less than 1s to compute for all networks; therefore, we omitted it from the table. Table IV also presents the mean time required for the computation of a sample-based approximation algorithm for betweenness and closeness centralities. We adopted two sample sizes for each network: 2.5% and 5% of the number of total vertices. The samples were uniformly randomized and five independent trials with distinct random seeds were executed for each sample size for each network.

The computation of the approximation algorithm was run in the same machine, but in a sequential environment. However, the computation of both algorithms shared similar parts; therefore, their computation was simultaneous for performance improvement. The times presented in Table IV comprise the computation of both metrics (betweenness and closeness).

The parallelization of the approximation algorithms is inefficient because it requires a larger number of dependent variables between different threads and a larger amount of memory. Parallelization reduced the execution time in about half but used 24 cores and 24 times more memory in the parallel environment. Notice that eigenvector and degree centralities are feasible even for massive networks, requiring orders of time less to compute than betweenness and closeness centralities, even though the computation of the previous were sequential and the computation of the latter made use of 24 cores in a parallel environment.

We also tested a sequential version for the latter centralities in the smaller networks (up to a hundred thousand vertices) used in the tutorial experiments. Experiments took around 16 times more computation time than their parallel versions, which indicates that the parallel version of the algorithm granted only about 2/3 of gain despite all threads being fully independent.

One can also observe the overhead effect of the sampling algorithms. Such algorithms maintain the same asymptotic time complexity upper bound, but at the expense of higher constant variables. For such reason, they are only viable when the size of networks compensates this effect, which, in the tests, occurred only for the networks with more than one million vertices when compared with the parallelized version of the exact algorithms.

We compared the results of the approximation algorithms (sample sizes of 2.5% and 5.0%) with the exact metrics using the Kendall τ-b correlation coefficient, which is a nonparametric measure of strength and association (interval [-1,1]) that exists between two variables measured on at least an ordinal scale. The 5% sample achieved a correlation of 0.9548, while the 2.5% sample achieved 0.9461. Therefore, we needed just a small fraction of vertices to approximate the centrality of all vertices of the network with high accuracy. However, the larger sample size, which required twice the time to compute, did not compensate its cost.

TABLE III. EXPERIMENTAL DATA: REAL NETWORKS DESCRIPTION

| Network | Type | Vertices | Edges | LCC % |
|---|---|---|---|---|
| *Stanford Large Network Dataset Collection* | | | | |
| Autonomous Systems AS-733 [82] | Routers | 6,474 | 12,572 | 100.00 |
| Oregon-1 [82] | Routers | 11,174 | 23,409 | 100.00 |
| Oregon-2 [82] | Routers | 11,461 | 32,730 | 100.00 |
| Astro Physics [83] | Collaboration | 18,772 | 196,972 | 95.37 |
| Condensed Matter [83] | Collaboration | 23,133 | 91,286 | 92.35 |
| General Relativity [83] | Collaboration | 5,242 | 13,422 | 79.32 |
| High Energy Physics [83] | Collaboration | 12,008 | 117,619 | 93.30 |
| High Energy Physics Theory [83] | Collaboration | 9,877 | 24,806 | 87.46 |
| Amazon [84] | Co-Purchasing | 334,863 | 925,872 | 100.00 |
| DBLP [84] | Collaboration | 317,080 | 1,049,866 | 100.00 |
| Youtube [84] | Social | 1,134,890 | 2,987,624 | 100.00 |
| Brightkite [85] | Social | 58,228 | 212,945 | 97.44 |
| Gowalla [85] | Social | 196,591 | 950,327 | 100.00 |
| Enron [86][87] | Email | 36,692 | 180,811 | 91.83 |
| Texas [86] | Road | 1,921,660 | 1,879,202 | 70.31 |
| Facebook [39] | Social | 4,039 | 82,143 | 100.00 |
| *Social Computing Data Repository* | | | | |
| Blog Catalog 3 [88] | Social | 10,312 | 333,983 | 100.00 |
| Buzznet [88] | Social | 101,168 | 2,763,066 | 100.00 |
| Delicious [88] | Social | 536,108 | 1,365,961 | 100.00 |
| Douban [88] | Social | 154,907 | 327,162 | 100.00 |
| Foursquare [88] | Social | 639,014 | 3,214,986 | 100.00 |
| Hyves [88] | Social | 1,402,611 | 2,184,244 | 100.00 |
| Livemocha [88] | Social | 104,438 | 2,193,083 | 99.68 |

| Network | Type | Vertices | Edges | LCC % |
|---|---|---|---|---|
| *BGU Social Networks Security Research Group* | | | | |
| The Marker Café [89][90] | Social | 69,411 | 1,644,781 | 99.86 |
| *The Koblenz Network Collection* | | | | |
| US Power Grid [72][91] | Supply Lines | 4,941 | 6,594 | 100.00 |
| Catster [91] | Social | 149,700 | 5,447,465 | 99.42 |
| Dogster [91] | Social | 426,820 | 8,543,322 | 99.92 |
| Hamster [91] | Social | 2,426 | 16,098 | 82.44 |
| Euroroad [91][92] | Road | 1,174 | 1,305 | 88.50 |
| Pretty Good Privacy [91][93] | Communication | 10,680 | 24,316 | 100.00 |

TABLE IV. METRICS COMPUTATION TIME WITH THE REAL NETWORKS.
DEGREE CENTRALITY REQUIRED LESS THAN 1 SECOND TO COMPUTE (AND IS OMITTED)

| Network | Time (hh:mm:ss) | | | |
|---|---|---|---|---|
| | Exact | | Sample 5% | Sample 2.5% |
| | $C_E$ | $C_B$ and $C_C$ | $C_B$ and $C_C$ | $C_B$ and $C_C$ |
| Euroroad | < 1s | < 1s | 00:00:52 | 00:00:23 |
| Hamster | < 1s | 00:00:03 | 00:01:42 | 00:00:49 |
| Facebook | < 1s | 00:00:05 | 00:03:31 | 00:01:42 |
| General Relativity | < 1s | 00:00:05 | 00:03:29 | 00:01:43 |
| US Power Grid | < 1s | 00:00:05 | 00:04:13 | 00:02:02 |
| Autonomous Systems AS-733 | < 1s | 00:00:11 | 00:05:35 | 00:02:45 |
| High Energy Physics Theory | < 1s | 00:00:25 | 00:07:21 | 00:03:36 |
| Pretty Good Privacy | < 1s | 00:00:25 | 00:09:07 | 00:04:33 |
| Oregon-1 | < 1s | 00:00:36 | 00:09:32 | 00:04:46 |
| Oregon-2 | < 1s | 00:00:37 | 00:09:50 | 00:04:49 |
| High Energy Physics | < 1s | 00:01:25 | 00:09:42 | 00:04:48 |
| Condensed Matter | < 1s | 00:03:20 | 00:18:54 | 00:09:23 |
| Blog Catalog 3 | 00:00:04 | 00:03:57 | 00:09:42 | 00:04:48 |
| Astro Physics | 00:00:01 | 00:04:42 | 00:16:19 | 00:08:00 |
| Enron | 00:00:01 | 00:10:28 | 00:30:46 | 00:15:12 |
| Brightkite | 00:00:02 | 00:25:46 | 00:53:41 | 00:26:26 |
| Douban | 00:00:03 | 02:10:51 | 02:45:48 | 01:23:02 |
| The Marker Cafe | 00:00:29 | 03:05:17 | 01:50:32 | 00:54:31 |
| DBLP | 00:00:09 | 03:33:41 | 07:40:50 | 03:49:58 |
| Buzznet | 00:00:47 | 03:41:01 | 03:10:24 | 01:35:47 |
| Amazon | 00:00:16 | 04:19:25 | 08:45:19 | 04:23:10 |
| Livemocha | 00:00:39 | 04:30:08 | 03:19:42 | 01:40:25 |
| Gowalla | 00:00:10 | 04:32:46 | 04:31:08 | 02:13:08 |
| Catster | 00:01:17 | 06:16:43 | 07:04:34 | 03:15:08 |
| Delicious | 00:00:20 | 09:47:57 | 15:34:38 | 07:53:01 |
| Foursquare | 00:00:44 | 24:23:38 | 32:33:36 | 16:04:30 |
| Texas | 00:00:10 | 33:38:20 | 49:06:11 | 24:07:02 |
| Dogster | 00:01:39 | 120:30:45 | 27:05:10 | 17:11:20 |
| Youtube | 00:01:03 | 188:07:33 | 44:04:45 | 22:28:10 |
| Hyves | 00:00:41 | 213:16:27 | 54:08:52 | 26:24:56 |

Next, we explain a faster (linear bounded) approximation technique for centrality measures that uses machine learning methods.

## 4.3 Machine Learning Methods: Applications, Experiments, and their Analyses

Some authors [57] [58] [94] have tried and experimented with approximation methodologies based mainly on machine learning and neural artificial networks. However, some of these works ([57] [58]) are limited because they only applied and showed the use of the technique for small networks with sizes up to 500 vertices, or focused the application to the approximation of eigenvector and PageRank centrality measures, which are feasible even for massive networks. In addition, none of the above works tried to optimize the neural network meta-parameters and structure using a generic neural network model instead. Here, we propose and explain a neural model methodology that has two major strengths: its *adaptability* (they can be used to approximate several distinct centrality measures) and its efficient computation time (they are able to compute the centrality measures a lot faster than any other method after the model has been trained). Moreover, we optimize the model considering several combinations of parameters and test it on real world networks comprising several thousands of vertices.

The methodology underlying the machine learning method is divided into four main steps.
 (i) First, the training data is acquired by using a complex network model to generate synthetic networks in which the centrality measures are computed. This is the data used for training the artificial network;
 (ii) Second, the training algorithm, network size and meta-parameters are selected for training the artificial neural network;
 (iii) Third, the accuracy of the model generated with the artificial neural network is compared with models generated by other machine learning techniques;
 (iv) Finally, the regression model is generated with the artificial neural network and applied to real-world networks in an application.

These steps are illustrated to facilitate the methodology understanding and reproduction of experiments using the techniques described here. In Fig. 2, the steps are used to illustrate how one can generate regression tasks in several applications. Each one of these steps will be detailed in the following sections.

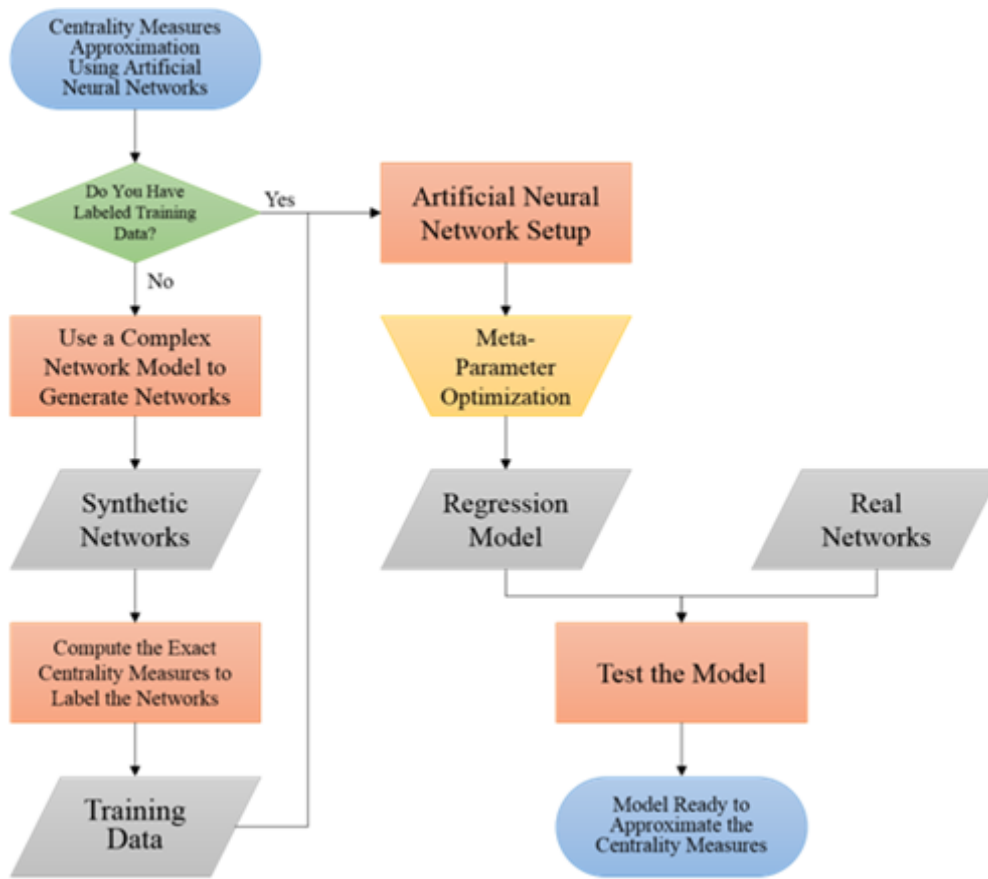

Fig. 2. Summary of the machine learning-based method for centrality measures.

## 4.4 Training Data Acquisition

When using any supervised machine learning method/algorithms, one important step is to obtain enough and consistent training data and to select the appropriate input data for the model. Since obtaining data from real networks is a costly process and that the computation of the centrality measures in huge networks is very expensive, one can generate one's own networks for such purposes using a complex network model.

The complex network model enables one to generate many synthetic networks with the properties enjoyed by real networks. Moreover, it allows one to generate smaller networks that reduce the computing costs to calculate the exact centrality measures, but – more importantly - keeping the most relevant structural properties presented by massive real-world networks.

The BTER complex network model was chosen for such tasks as it is one of the best models to reproduce real world networks structural properties; moreover, it is easy to implement and configure. BTER requires two configuration parameters: the desired degree distribution and the clustering coefficient (which can be configured as a global value, by vertices degree or by community structure). In the tutorial, we applied both a heavy-tailed and a lognormal distribution as degree distribution to provide generic data to train and learn the model. Both distributions are known as the best representatives of most real networks degree distribution studied in the literature [95] [96].

Table V [57] summarizes the formula of each function used to model the degree distribution and the parameter values used in the experiments. The values generated by each function were considered as proportions (weights) that a given degree appears in a given network size [57].

TABLE V. DEGREE DISTRIBUTIONS OF VERTICES

| Distribution | Formula | Parameters |
|---|---|---|
| Heavy-tailed | $k^{-\lambda}$ | $\lambda = \{1.5, 2, 2.5\}$ |
| Log-normal | $e^{\frac{-(\ln k)^2}{S}}$ | $S = \{5, 10, 15\}$ |

Source [57]

The global clustering coefficient of many real networks belongs to the real numbers interval between 0.1 and 0.5, but preliminary tests with the BTER led to higher choices of clustering coefficients as parameter of the model [57]. This is mainly because the parameter set of the model is the maximum clustering coefficient to which the network generated [57]. Considering this, we selected as desired clustering coefficients for the model random values in the real numbers interval [0.3, 0.7] [57].

There are also other configurable parameters to consider during the construction of the networks, like degree-1 vertices and groups of vertices with mixed degree [57]. Those are selected following the suggestions made by the BTER authors [11].

The training data contained 10 networks for each kind of degree distribution (Table V) with sizes ranging from 100 to 1,000 vertices, totaling 600 synthetic networks and 330 thousand vertices. These relatively small sizes were selected to enable the computation of the exact ranks of the metrics (betweenness and closeness) used as labels during the training.

The second issue one has to deal with is the selection of the proper input attributes to train the model. Each centrality measure uses complex properties of the graph representing the network and the computation of most of these properties are the main reason that centrality measures are time expensive. For such reason, in the tutorial we selected the fastest centrality measures (degree and eigenvector), which are computationally feasible even for massive networks and are highly related to the other metrics [12]. This provides generality to this tutorial as such metrics summarize most information needed to compute the other extant metrics.

Therefore, we compute degree and eigenvector to serve as input data and we chose to compute the exact values of betweenness and closeness as desired output values (supervised learning) in the networks generated with the BTER model. Closeness and betweenness metrics were chosen because they are the most applied centrality measures but, at the same time, their exact algorithms computation is unfeasible or require too much time to compute in massive networks and dynamic networks. The increasing availability and interest of study on such networks created an urge for approximation techniques for both measures, as they are important analysis tools for many areas [53].

We computed the rank of each vertex in each centrality measure, as we are interested in the order of such vertices and not in the absolute centrality values. Consequently, a sample input of training data will contain the rank of a vertex considering the eigenvector and degree centralities (2 inputs) and the outputs/label/desired values will be the rank of the same vertex in one of the other centralities (betweenness and closeness in our experiments).

In addition, when one is considering the use of the same technique to approximate other centrality measures (e.g. walk betweenness, hierarchical closeness or subgraph centralities), the input information will remain the same for the simplest centrality measures (degree and eigenvector). The desired measure/s need to be computed in the synthetic networks to serve as output labels during the training. Notice that such computation is feasible due to the use of the complex network model that generates smaller synthetic networks with properties similar to that of the massive networks that will later be used in the application environment.

Both input and desired values are first normalized by the size of the network, then to belong to the interval [-1, 1] and finally to have zero mean and variance equal to one. The preprocessing helps configure the artificial neural networks in an easier way and allows for faster training.

## 4.5 Artificial Neural Network Training

The appropriate training of an artificial neural network can be a complex issue. Many techniques and methodologies, algorithms and tools can be applied in specific tasks. For such reason, we applied the neural network toolbox from MATLAB 2015 to configure and train the neural networks. This toolbox comprises most of the algorithms that are well established in the field with a built-in parallelism pool and a robust implementation for industrial, academic and research use.

Before one starts the training, one needs to select several meta-parameters for an artificial (neural) network, such as size and structure of the network, learning algorithm, and learning parameters. Therefore, in the experiments, we initially optimized these parameters using 10-fold cross-validation to find out the best configuration to approximate the centrality measures.

The batch method is typically used to update the weights of the network. In batch training, the adjustment delta values are accumulated over all training items to give an aggregate set of deltas and then they are applied to each weight and bias.

In the experiments, we stopped the training after 5min to check the fastest training configuration, so to avoid the use of a large unnecessary number of parameters and to prevent overfitting. Our objective here was to find out the most efficient configuration of parameters considering both its computational costs and solution quality. The quality of the solution was measured by the determination coefficient ($R^2$) using only the test set results, which considers 10% of total data.

We have selected the fully-connected feedforward multilayer perceptron neural network architecture as first approach as this model is robust to outliers, generates a highly flexible model and it is simple to implement and train using the backpropagation learning algorithm.

First, we experimented with several networks sizes (numbers of neurons and hidden layers). The number of neurons of the networks with more than one hidden layer was selected in a way that they match the total number of learning parameters of the network with only one hidden layer. For example, the neural network with a single hidden layer with 175 neurons comprises 525 learning parameters in our configuration, the same exact number of parameters presented by a two-hidden layer network with 21 neurons in each layer and of a three-hidden layer network with 15 neurons in each layer. The same was valid for each of the other groups of test cases (7-3-2, 11-4-3, 25-7-5, 56-11-8, 84-14-10, 299-28-20 and 532-38-27 respectively to one hidden layer, two hidden layers, and three hidden layers). The total number of learning parameters ranged from 18 to 1596.

In this way, one can make a direct comparison of the effect caused by the number of layers and discover the most efficient configuration of parameters. Notice that the more parameters the network comprises, the better is its performance, up to a tipping point where the complexity of its training computation does not pay off and training efficiency decays. Therefore, an important objective is also to identify such tipping point to optimize the learning performance.

One can select the training algorithm suggested by MATLAB environment as default (the Levenberg-Marquardt algorithm) to train the neural networks in this stage. Additionally, we set the activation function of all hidden layers to hyperbolic tangent (any sigmoid function would suffice for such task) and the activation function of the output layer as a linear function (due to the regression task in hand).

In the experiments (Fig. 3), a three-hidden layer network with 20 neurons in each layer was the architecture that presented the best trade-off between solution quality and training speed. The larger networks suffered from overfitting and complexity demands, while smaller networks hindered solution quality.

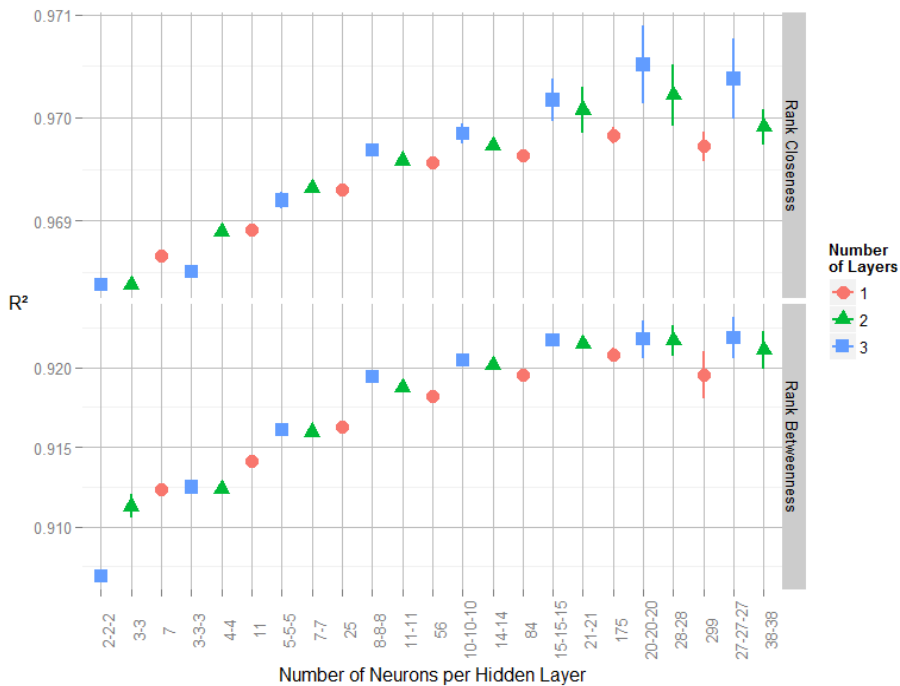

Fig. 3. Outline of results for different artificial neural network structures. Summary of results for each centrality measure ranked with 99% confidence intervals (i.e. the expected mean value is within the shown interval with 99% chance if the experiment is repeated). When the interval does not appear in the figure, it means that the variance is too small to be significant in this scale.

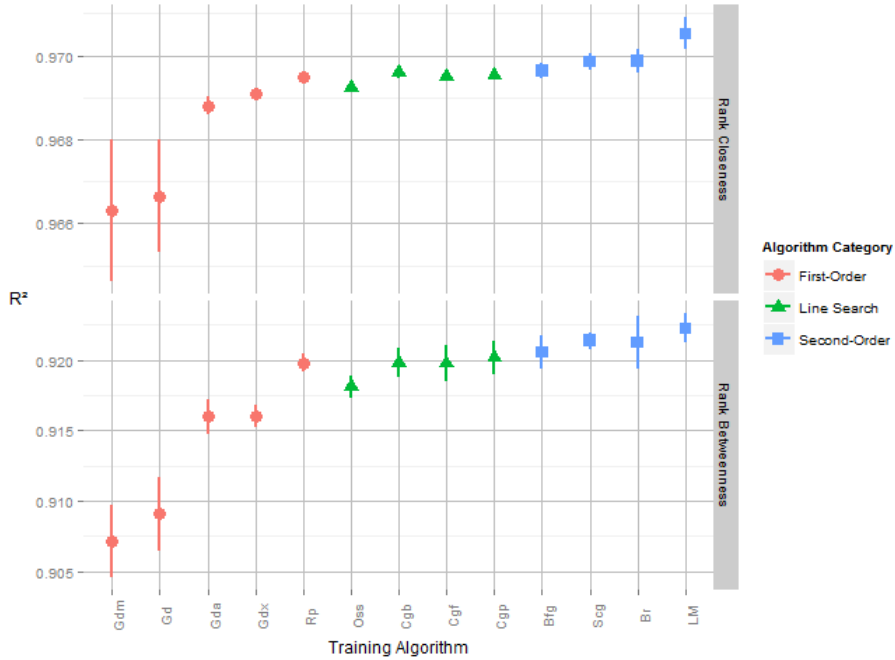

Fig. 4. Summary of results for the artificial neural network algorithms, for each centrality measure; ranked version with 99% confidence intervals. The algorithms were grouped by their category, and then by increasing order of solution quality. We depict only the results for the three-hidden layer network architecture with 20 neurons in each layer – all the other architectures exhibited a similar pattern.

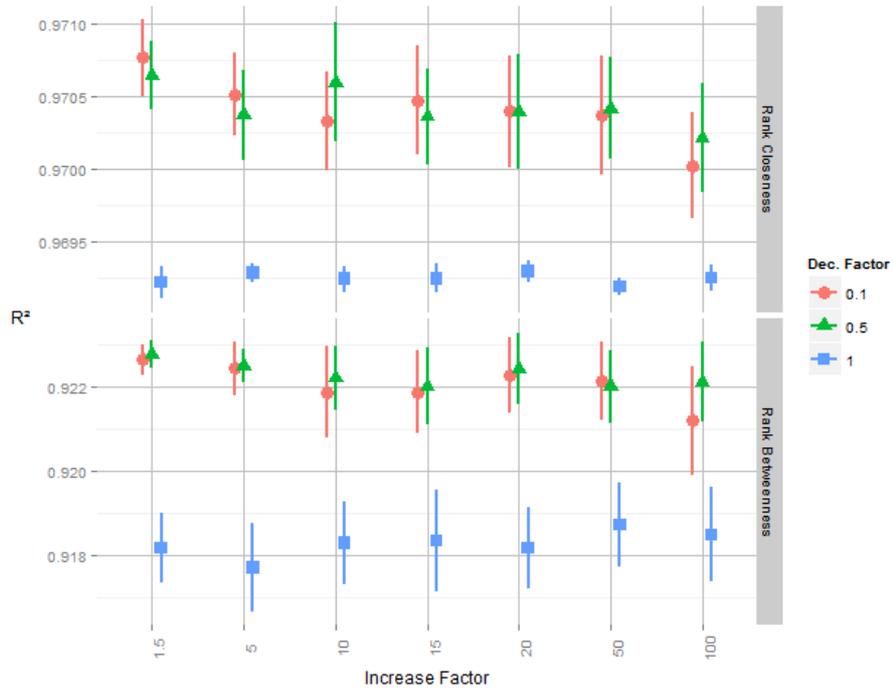

Fig. 5. Outline of results for each combination of the LM algorithm parameters, for each centrality measure; ranked version with 99% confidence intervals.

In the next stage of training, we experimented with all the backpropagation algorithms available in MATLAB's neural network toolbox for feedforward networks (notice that, of course, many freely available machine learning software tools offer several versions of such algorithms). We tested all the combinations of previously selected network architectures with all algorithms. Table VI [57] depicts the algorithms and their selected parameter values for this experimental stage. The parameters are the ones selected by default by the MATLAB environment and are more robust for a large number of applications. The fact is noticeable in Fig. 4 where the performance of the different algorithms is depicted.

One can notice that in Fig. 4, although the difference between most algorithms is considerably small, the second-order algorithms perform better than the ones with line search, which in their turn were generally better than the first-order algorithms. Moreover, the LM algorithm presents a slightly better overall result than all others do for Rank Closeness despite being statistically like all second-order algorithms for Rank Betweenness. Therefore, the LM algorithm was selected to further improvements and experiments in the final stage of parameter optimizations.

In the third and final stage, we optimize the parameters within the LM method. The Marquardt adjustment (*mu*) initial value is set to a value close to zero. This allows the training algorithm to apply larger weight updates and so speed up the initial convergence, due to the fact that the weight values are initially set randomly (with lower and maximum bounds to avoid an initial and detrimental saturation of the activation function). It is not very important what exact value is chosen considering that the algorithm will further adjust the size of the steps dynamically during the training with two other parameters: *mu* decrease and increase factors. The LM method has also a parameter to set a maximum value for *mu*, which limits the method to a smallest step size and can avoid overfitting and the waste of time with finer grained, but nearly useless steps. In our application, this parameter can be set to an arbitrarily high value since we also use a maximum time limit to stop the training early and later (in the final setup and training) we use a validation set to avoid overfitting.

Thus, we focus on the optimization of the *mu* decrease/increase factors, which are responsible for adjusting the *mu* factor at each training iteration and are usually sensible to each specific application, requiring finer adjustments to achieve the best results in solution quality and training performance. The decrease factor parameter is a proportion of the increase factor, preferably smaller than the other to prevent loops during training. The exact number of the increase and decrease factors depends on the numerical amplitude of the training data. As they are not easy to estimate, we tested a wide range [1.5,100] of parameter combinations. The results of this comparison can be checked in Fig 5.

We can note by looking at Fig. 5 that excluding the decreasing factor value of 1 (100% of the increase factor), all other combinations of parameters were statistically similar within the confidence intervals. This fact demonstrates that these parameters do not interfere as much as one may think in this task. However, the combination 1.5 with 0.1 of increasing and decreasing factors, respectively, presents the lowest variance and the highest mean in our experimental results. Hence, we trained our final model using a fully connected artificial three-hidden layer network with 20 neurons in each layer (Fig. 6) and the LM algorithm with initial/maximum mu set to $0.005/10^{10}$ and increasing/decreasing mu factor set to 1.5/0.1.

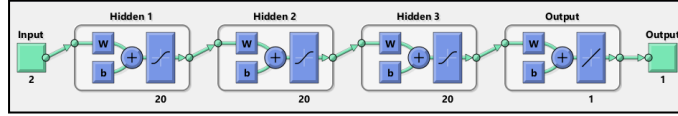

Fig. 6. Artificial neural network architecture for the experiments with the LM algorithm.

TABLE VI. PARAMETERS OF THE BACKPROPAGATION ALGORITHM

| Algorithm | Parameters |
|---|---|
| Gradient Descent (Gd) [97] | Learning rate = 0.01 |
| Gradient Descent with Momentum (Gdm) [97] | Learning rate = 0.01<br>Momentum constant = 0.9 |
| Variable Learning Rate Gradient Descent (Gdx) [97] | Learning rate = 0.01<br>Momentum constant = 0.9<br>Increase/decrease ratio to learning rate = 1.05/0.7 |
| One-Step Secant (Oss) [98] | Linear search = 1-D minimization backtracking [99]<br>Initial step size = 0.01<br>Scale factor to sufficient performance reduction = 0.001<br>Scale factor that determine sufficiently large step size = 0.1<br>Step size lower/upper limit = 0.1/0.5<br>Minimum/maximum step length = $10^{-6}/100$<br>Linear search tolerance = 0.0005 |
| BFGS Quasi-Newton (Bfg) [100] | The same parameters and values as the method above. |
| Polak-Ribiére Conjugate Gradient (Cgp) [101] | Linear search = 1-D minimization using Charalambous' method [102]<br>Initial step size = 0.01<br>Scale factor to sufficient performance reduction = 0.001<br>Scale factor that determine sufficiently large step size = 0.1<br>Scale factor to avoid small performance reductions = 0.1<br>Linear search tolerance = 0.0005 |
| Fletcher-Powell Conjugate Gradient (Cgf) [101] | The same parameters and values as the method above. |
| Conjugate Gradient with Powel/Beale Restarts (Cgb) [103] | The same parameters and values as the method above. |
| Scaled Conjugate Gradient (Scg) [104] | Change in weight for the second derivative approximation = $5.10^{-5}$<br>Regulation of the Hessian indefiniteness = $5.10^{-7}$ |
| Resilient Backpropagation (Rp) [105] | Learning rate = 0.01<br>Initial weight change = 0.07<br>Increment/decrement to weight change = 1.2/0.5<br>Maximum weight change = 50 |
| Bayesian Regularization (Br) [106] | Marquardt adjustment (*mu*) = 0.005<br>*mu* decrease/increase factor = 0.1/10<br>*mu* maximum value = $10^{10}$ |
| Levenberg-Marquardt (LM) [107] | The same parameters and values as the method above. |

Source [57]

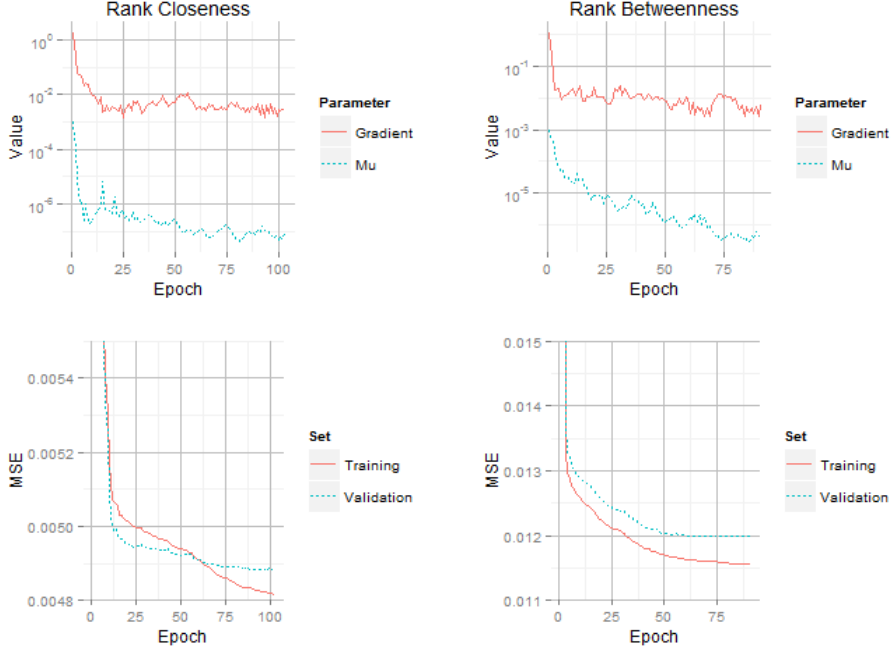

Fig. 7. Training behavior of the parameters and the mean squared error (MSE) evolution over each epoch of training for each target centrality measure.

Finally, we tested a different input configuration with the addition of two-hop degree rank (a vertex degree value summed with the degree of its immediate neighbors). In this experiment we only tested the neural network configuration for the final model. The results showed that the addition of a new input improved the accuracy of the model to approximate closeness centrality by 5% on average, but it reduced the accuracy for betweenness centrality by 10% on average. Therefore, we chose to use just the two basic inputs (degree and eigenvector ranks) for the final training of the model.

For the setup that generates the final model, we divided the data uniformly at random (330,000 vertices from 600 different synthetic networks) into a set with 85% for training and a set with 15% for validation to prevent overfitting. The training stops whenever the solution does not improve for the validation set in 10 consecutive full batches. We set no training time limit (Fig. 7).

Notice that although MSE continues to drop, the validation set serves as an early stopping criterion to prevent overfitting, because it was not used for training. The training took about 10min for rank closeness and about 7min for rank betweenness. We run the experiments in the same machine specifications with all CPU cores allocated for parallelism. The machine had the following specifications: Intel Core i7-5820K processor with six 3.3Ghz physical cores with 15MB shared cache memory, quad-channel 4x4GB DDR4 2133MHz RAM memory, Windows 10 operating system.

The learning/regression of rank betweenness showed higher difficulty than rank closeness, even though both presented very low error bounds. This is supported by the final MSE values, one of each is half the other, but both are considerably small, since the output is in the interval [-1,1] and we used 330 thousand samples for training.

### 4.6 Comparison Between Different Machine Learning Models

There are many machine learning techniques capable of creating regression models in tasks such as the ones tackled in this tutorial. To reinforce the application of neural learning models we compared their performance with other machine learning techniques from the literature and also available in the MATLAB environment. For such purpose we applied the same training data and configuration applied for the neural learning algorithm (depicted in Subsections 4.4 and 4.5).

We applied a 10-fold cross-validation analysis to compute the $R^2$ as comparison factor. Notice that due to the high number of samples in the training data even small differences in the $R^2$ means a considerable disparity in the performance. The results and the algorithms applied in our experiments are described in Table VII. Due to the robustness of the implementation of the algorithms in MATLAB, all the 99% confidence intervals lie in the fourth decimal place; therefore, they are not shown in the Table.

TABLE VII. MACHINE LEARNING MODELS COMPARISON

| Learning Algorithm | Description | Interpretability | Flexibility | $C_C$ $R^2$ | $C_B$ $R^2$ |
|---|---|---|---|---|---|
| Linear Regression | A linear regression model with only intercept and linear terms | Easy | Very low | 0.95 | 0.87 |
| Interactions Linear | A linear regression model with intercept, linear and interaction terms | Easy | Medium | 0.95 | 0.87 |
| Robust Linear | A robust (less sensitive to outliers) linear regression model with only intercept and linear terms | Easy | Very low | 0.95 | 0.86 |

| Learning Algorithm | Description | Interpretability | Flexibility | $C_C$ $R^2$ | $C_B$ $R^2$ |
|---|---|---|---|---|---|
| Stepwise Linear | A linear model with terms determined by a stepwise algorithm | Easy | Medium | 0.95 | 0.87 |
| Fine Tree | A fine regression tree with minimum leaf size of 4 | Easy | High | 0.96 | 0.88 |
| Medium Tree | A medium regression tree with minimum leaf size of 12 | Easy | Medium | 0.96 | 0.89 |
| Coarse Tree | A coarse regression tree with minimum leaf size of 36 | Easy | Low | 0.95 | 0.89 |
| Boosted Trees | An ensemble of regression trees using the LSBoost algorithm | Hard | Medium to High | 0.95 | 0.87 |
| Bagged Trees | A bootstrap-aggregated ensemble of regression trees | Hard | High | 0.95 | 0.88 |
| Linear SVM | A support vector that follows simple linear structure in the data, using the linear kernel | Easy | Low | 0.95 | 0.87 |
| Quadratic SVM | A support vector machine that uses the quadratic kernel | Hard | Medium | 0.96 | 0.87 |
| Fine Gaussian SVM | A support vector machine that follows finely-detailed structure in the data. It uses the Gaussian kernel with kernel scale $\sqrt{1/2}$ | Hard | High | 0.96 | 0.88 |
| Medium Gaussian SVM | A support vector machine that finds less fine structure in the data. It uses the Gaussian kernel with kernel scale $\sqrt{2}$ | Hard | Medium | 0.96 | 0.87 |
| Coarse Gaussian SVM | A support vector machine that follows coarse structure in the data. It uses the Gaussian kernel with kernel scale $4\sqrt{2}$ | Hard | Low | 0.96 | 0.87 |
| MLP Neural Network with Backpropagation | A Multilayer Perceptron Neural Network Implementation with Backpropagation Learning | Hard | High | 0.97 | 0.92 |

We can check in the Table that the neural network architecture performs considerably better than all other techniques to approximate Betweenness centrality and slightly better for Closeness centrality. It proved to be the more flexible while robust tested methodology in our experiments although it may be hard to interpret.

**4.7 Artificial Neural Network Learning with Real World Network Data**

The final stage of a machine learning application is testing the respective learning algorithm/model with real world data to validate the model. In the tutorial experiments, we used 30 real-world networks from four freely available data repositories of network data. All the selected networks were symmetric and binary (unweighted edges). The largest connected component (LCC) is used in the experimental validation, and LCC was computed for all analyzed networks.

First, we computed eigenvector and degree centralities for all vertices of the real networks. The rank of the vertices in each network considering each of the centralities is then used as inputs for the machine learning model to approximate its rank in a target selected centrality (betweenness or closeness in our experiments). Notice that a different model is used depending on the target centrality measure because a different model was trained specifically to approximate each centrality (Section 4.5).

We also computed the exact values for the betweenness and closeness centralities for all the networks, but they are only used to compute the precision/error of the results provided by the approximation methods (Figs. 8 and 9) and to compare their cost in time (Table IV of Section 4.2).

The next stage is to run the model generated by the artificial neural network and previously trained to approximate betweenness and closeness for the vertices of the real networks. In order to do so, we used the same computer configuration of the training tasks in a parallel environment with 6 cores. The computation took less than 1s for any of the networks, which is a significant result and illustrates the effectiveness of the machine learning methodology.

We then compared the results generated by the neural network (NN) with the results of the approximation algorithms (with sample sizes of 2.5% and 5.0%) using the Kendall τ-b correlation coefficient.

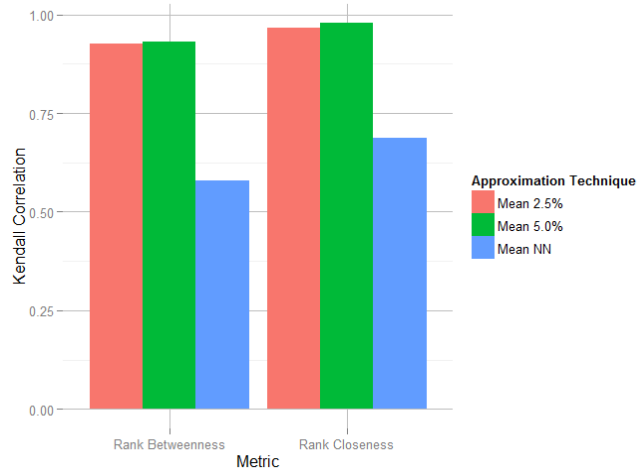

Fig. 8. Mean correlation coefficients computed for all real-world networks.

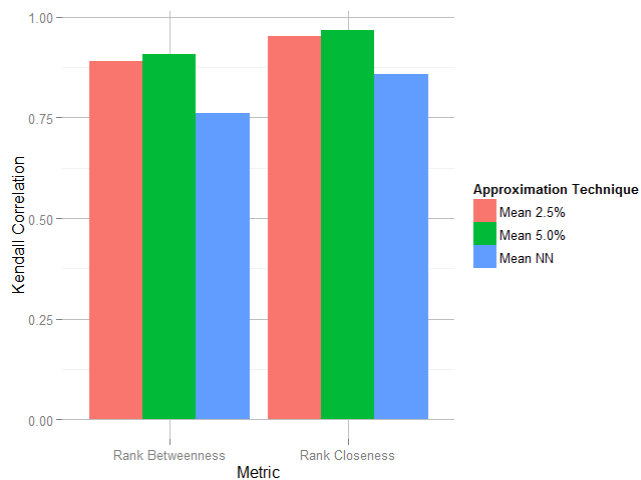

Fig. 9. Mean correlation coefficients considering only three networks (Blog Catalog 3, Foursquare and Livemocha). These were the top 3 networks w.r.t. the model's performance for both centrality measures.

One can see that there is just a small gain for the approximation algorithms when the size of sample is doubled (notice the difference between 2.5% and 5.0% above), which is hardly worth the cost considering that the computation time is doubled. On the other hand, the results obtained with the model were lower in quality, but still granted good results (over 0.55 for betweenness and 0.65 for closeness centralities). Moreover, because of the generic formulations based mainly on social networks characteristics that were input for the BTER model (which generates the synthetic networks used for the artificial neural network training), we expected a large variance in the results presented by the model considering distinct kinds of networks. This, however, is not observed for the sample-based algorithms, which perform nearly equally well in all networks tested. One can see that the learning model achieved a poorer result for the infrastructure networks (Euroroad, Texas, and US Power Grid for instance) and a better performance in social networks.

These results show that even though the neural network model was not capable to rank all the vertices in the correct order, it is effective enough to classify chunks of vertices as highly central and has a close performance with respect to the sample approximation methods that require considerably more computational time. Many applications of centrality measures are interested in the selection of the top central vertices and not in their exact centrality order. For instance, the selection of influencers to optimize advertisement, the control and protection of critical spots in a communication network and the allocation of resources to enhance their distribution usually make use of a group of elements and do not rely on their specific order. Moreover, these kinds of networks, massive and constantly evolving require a continuous analysis and computation of such metrics. Therefore, the time saved by the use of a machine learning model is quite relevant in such applications.

However, the use of machine learning may not be adequate to approximate centralities in applications that rely on the exact centrality values or are oversensitive to the order of the centrality rank. They are more adequate for applications where the use of centrality measures offers complementary analysis or are used as part of a decision process or heuristics.

## 5 CONCLUSIONS

The growing rate of network research and applications demands the development of principled research models and tools for network analysis. Centrality measures are widely used in many application domains. However, as networks grow in size, their computation costs present several challenges, especially when working with complex real-world networks. These challenges may hinder some important applications and studies.

Machine learning techniques have recently been successful in a number of relevant applications that manage large amounts of data [75] [107] [108] [109]. Moreover, the referred growing availability of massive amounts of network data demands the use of effective machine learning techniques to better exploit and to interpret these data. In this context, we presented a tutorial where we have explained how one can propose, explain, test, and compare centrality measures using artificial neural networks learning. To do so, we have identified the best configuration for the artificial neural network training, including network structure, training algorithm, and training meta-parameters.

We have shown how one can use a generative model - such as the BTER complex network model - as a way to provide unlimited training data as it may be configured to generate synthetic networks of any size and with similar structural patterns for any kind of application. The experimental results revealed that two simple centralities, degree and eigenvector, can be used as input vector to approximate closeness and betweenness centralities. Experimental results show us that the machine learning methodology can be used with any other similar metric as targeted value, which increases their potential use.

The model is able to approximate the target centrality measures with considerable accuracy and reduce computation costs in 30 real world experimental case scenarios. The model showed an incredible advantage and tradeoff with respect to computation costs, making it a viable option for applications where accuracy is not the fundamental goal and the computation resources are limited, but where approximations via machine learning are effective alternatives.

Suggested future research avenues include the use and development of multi-tasking artificial neural networks capable of learning multiple centrality measures at once, and the approximation of temporal centrality measures using recurrent neural models. The development of tools capable of automatically generating/training artificial neural networks, which approximate several centrality measures for a given generic network, would also facilitate further applications. The methodology presented in this tutorial paper elicits the requirements towards building such tools.

## REFERENCES

[1] D. Easley and J. Kleinberg. *Networks, Crowds, and Markets: Reasoning About a Highly Connected World*. Cambridge University Press, 2010.
[2] David Liben-Nowell, Jon M. Kleinberg. The link-prediction problem for social networks. *JASIST* 58(7): 1019-1031, 2007.
[3] Yiling Chen, Arpita Ghosh, Michael Kearns, Tim Roughgarden, Jennifer Wortman Vaughan: Mathematical foundations for social computing. *Commun. ACM* 59(12): 102-108, 2016.
[4] Shui Yu, Meng Liu, Wanchun Dou, Xiting Liu, Sanming Zhou: Networking for Big Data: A Survey. *IEEE Communication Surveys & Tutorials*, 19(1): 531-549, 2017.
[5] E. Bastug, K. Hamidouche, W. Saad and M. Debbah Centrality-based caching for mobile wireless networks, "*1st KuVS Workshop on Anticipatory Networks*", 1-4, 2014.
[6] E. Baştuğ, M. Bennis and M. Debbah Living on the edge: the role of proactive caching in 5G wireless networks, "*IEEE Communications Magazine*" 52(8):82-89, 2014.
[7] M. E. Shaw, "Group structure and the behavior of individuals in small groups", *Journal of Psychology*, 38:139-149, 1954.
[8] J. Nieminen, "On centrality in a graph", *Scandinavian Journal of Psychology*, 15: 322-336, 1974.
[9] L. C. Freeman, "Centrality in social networks: conceptual clarification", *Social Network*, 1:215-239, 1978/79.
[10] C. Seshadhri, T. G. Kolda, and A. Pinar, "Community structure and scale-free collections of Erdős-Rényi graphs", *Physical Review E*, 85(5):056109, 2012.
[11] T. G. Kolda, A. Pinar, T. Plantenga and C. Seshadhri, "A scalable generative graph model with community structure", *SIAM Journal on Scientific Computing*, 36(5):424-452, 2014.
[12] F. Grando, D. Noble, and L. C. Lamb, "An analysis of centrality measures for complex and social networks", *Proc. of IEEE Global Communications Conference*, 1-6, 2016.
[13] L. C. Freeman, "A set of measures of centrality based on betweenness", *Sociometry*, 40:35-41, 1977.
[14] U. Brandes, "A faster algorithm for betweenness centrality", *Journal of Mathematical Sociology*, 25:163-177, 2001.
[15] L. C. Freeman, S. P. Borgatti and D. R. White, "Centrality in valued graphs: a measure of betweenness based on network flow", *Social Networks*, 13(2):141-154, 1991.
[16] M. E. J. Newman, "A measure of betweenness centrality based on random walks", *Social Networks*, 27(1):39-54, 2005.
[17] E. Estrada, D. J. Higham, and N. Hatano, "Communicability betweenness in complex networks", *Physica A: Statistical Mechanics and its Applications*, 388(5):764-774, 2009.
[18] M. Ercsey-Ravasz, R. N. Lichtenwalter, N. V. Chawla and Z. Toroczkai, "Range-limited centrality measures in complex networks", *Physical Review E*, 85(6), 2012.
[19] L. Maccari, and R. L. Cigno, "Pop-routing: centrality-based tuning of control messages for faster route convergence", *Proc. of the 35th IEEE Internacional Conference on Computer Communications*, 1-9, 2016.
[20] A. Vázquez-Rodas and L. J. de la C. Llopis, "A centrality-based topology control protocol for wireless mesh networks", *Ad Hoc Networks*, 24(B):34-54, 2015.
[21] J. Ding and Y-Z. Lu, "Control backbone: an index for quantifying a node's importance for the network controllability", *Neurocomputing*, 153:309-318, 2015.
[22] L. Baldesi, L. Maccari and R. Lo Cigno, "Improving P2P streaming in wireless community networks", *Computer Networks*, 93(2):389-403, 2015.
[23] M. Kas, S. Appala, C. Wang, K. Carley, L. Carley and O. Tonguz, "What if wireless routers were social? Approaching wireless mesh networks from a social perspective", *IEEE Wireless Communications*, 19(6):36-43, 2012.
[24] P. Pantazopoulos, M. Karaliopoulos, and I. Stavrakakis, "Distributed placement of autonomic internet services", *IEEE Transactions on Parallel and Distributed Systems*, 25(7):11702-1712, 2014.
[25] L. Maccari and R. L. Cigno, "Waterwall: a cooperative, distributed firewall for wireless mesh networks", *EURASIP Journal on Wireless Communications and Networking*, 1:1-12, 2013.
[26] L. Maccari, Q. Nguyen, and R. L. Cigno, "On the computation of centrality metrics for network security in mesh networks", *Proc. of the Global Communications Conference*, 1-6, 2016.
[27] P. Zilberman, R. Puzis and Y. Elovici, "On network footprint of traffic inspection and filtering at global scrubbing centers", *IEEE Transactions on Dependable and Secure Computing*, PP(99):1-16, 2015.
[28] M. D. König and C. J. Tessone, "From assortative to dissortative networks: the role of capacity constraints", *Advances in Complex Systems*, 13(4):483-499, 2010.
[29] M. D. König, C. J. Tessone and Y. Zenou, "Nestedness in networks: a theoretical model and some applications", *Theoretical Economics*, 9:695-752, 2014.
[30] M. E. J. Newman, and J. Park, "Why social networks are different from other types of networks", *Physical Review E*, 68:036122, 2003.
[31] M. E. J. Newman, and M. Girvan, "Finding and evaluating community structure in networks", *Physical Review E*, 69(2):026113, 2004.

[32] X. Cao, L. Wang, B. Ning, Y. Yuan, and P. Yan, "Pedestrian detection in unseen scenes by dynamically updating visual words", *Neurocomputing*, 119:232-242, 2013.
[33] P. Kazieenko, T. Kajdanowicz, "Label-dependent node classification in the network", *Neurocomputing*, 75(1):199-209, 2012.
[34] Y. Xu, X. Hu, Y. Li, D. Li and M. Yang, "Using complex network effects for communication decisions in large multi-robot teams", *Proc. of the 13th International Conference on Autonomous Agents and Multiagent Systems*, 685-692, 2014.
[35] P. Moradi, M. Ebrahim, and N. Entezari, "Automatic skill acquisition in reinforcement learning agents using connection bridge centrality", *Communication and Networking*, 120:51-62, 2010.
[36] A. A. Rad, M. Hasler and P. Moradi, "Automatic skill acquisition in reinforcement learning using connection graph stability centrality", *Proc. of the 2010 IEEE International Symposium on Circuits and Systems*, 697-700, 2010.
[37] J. A. Danowski and N. Cepela, "Automatic mapping of social networks of actors from text Corpora: time series analysis", *Data Mining for Social Network Data*, 12:31-46, 2010.
[38] J. C. Jiang, J. Y. Yu and J. S. Lei, "Finding influential agent groups in complex multiagent software systems based on citation network analyses", *Advances in Engineering Software*, 79:57-69, 2015.
[39] J. Mcauley and J. Leskovec, "Learning to discover social circles in ego networks", *Proc. of the Advances in Neural Information Processing Systems*, 2012.
[40] E. Yan and Y. Ding, "Applying centrality measures to impact analysis: a coauthorship network analysis", *Journal of the Association for information Science and Technology*, 60(10):2107-2118, 2009.
[41] A. Louati, J. El Haddad and S. Pinson, "A multilevel agent-based approach for trustworthy service selection in social networks", *Proc. of the IEEE/WIC/ACM International Joint Conferences on Web Intelligence and Intelligent Agent Technologies*, 214-221, 2014.
[42] S. Kaza and H. Chen, "Identifying high-status vertices in knowledge networks", *Data Mining for Social Network Data*, 12:91-107, 2010.
[43] X. Li, Y. Liu, Y. Jiang, and X. Liu, "Identifying social influence in complex networks: a novel conductance eigenvector centrality model", *Neurocomputing*, 210:141-154, 2016.
[44] G. Hua, Y. Sun, and D. Haughton, "Network analysis of US Air transportation network", *Data Mining for Social Network Data*, 12:75-89, 2010.
[45] S. Gao, Y. Wang, Y. Gao and Y. Liu, "Understanding urban traffic-flow characteristics: a rethinking of betweenness centrality", *Environment and Planning B: Urban Analytics and City Science*, 40(1):135-153, 2013.
[46] A. Jayasinghe, K. Sano and H. Nishiuchi, "Explaining traffic flow patterns using centrality measures", *International Journal for Traffic and Transport Engineering*, 5(2):134-149, 2015.P. X. Zhao and S. M. Zhao, "Understanding urban traffic flow characteristics from the network centrality perspective at different granularities", *The International Archives of the Photogrammetry, Remote Sensing and Spatial Information Sciences*, XLI-B2:263-268, 2016.
[47] D. Noble, F. Grando, R. M. Araújo, and L. C. Lamb, "The impact of centrality on individual and collective performance in social problem-solving systems", *Genetic and Evolutionary Computation Conference*, 2015.
[48] T. P. Michalak, K. V. Aadithya, P. L. Szczepański, B. Ravindran, and N. R. Jennings, "Efficient computation of the Shapley value for game-theoretic network centrality", *Journal of Artificial Intelligence Research*, 46:607-650, 2013.
[49] M. O. Jackson and A. Watts, "The evolution of social and economic networks", *Journal of Economic Theory*, 106(2):265-295, 2002.
[50] W. Xiong, L. Xie, S. Zhou and J. Guan, "Active learning for protein function prediction in protein-protein interaction networks", *Neurocomputing*, 145:44-52, 2014.
[51] M. Rubinov and O. Sporns, "Complex network measures of brain connectivity: uses and interpretations", *NeuroImage*, 52(3):1059-1069, 2010.
[52] M. P. van den Heuvel and O. Sporns, "Network hubs in human brain", *Trends in Cognitive Sciences*, 17(12):683-696, 2013.
[53] K. Das, S. Samanta and M. Pal. "Study on centrality measures in social networks: a survey", *Social Network Analysis and Mining*, 1-11, 2018.
[54] A. Landherr, B. Friedl and J. Heidemann. "A critical review of centrality measures in social networks". *Business & Information Systems Engineering*, 2(6):371-385, 2010.
[55] L. F. Costa, F. A. Rodrigues, G. Travieso and P. R. Villas Boas, "Characterization of complex networks: a survey of measurements", *Advances in Physics*, 56:167-242, 2008.
[56] A. Bonato. "A survey of properties and models of online social networks". *International Conference on Mathematical and Computational Models*. 1-10, 2009.
[57] F. Grando and L. C. Lamb, "On approximating networks centrality measures via neural learning algorithms", Proc. of the International Joint Conference on Neural Networks (IJCNN), 551-557, 2016.
[58] F. Grando and L. C. Lamb, "Estimating complex networks centrality via neural networks and machine learning", Proc. of the International Joint Conference on Neural Networks (IJCNN), 1-8, 2015.
[59] A. Bavelas, "A mathematical model for group structures", *Human Organization*, 7:16-30, 1948.
[60] G. Sabidussi, "The centrality index of a graph", Psychometrika, 31:581-603, 1966.
[61] U. Brandes, "On variants of shortest-pat betweenness centrality and their generic computation", *Social Networks*, 30:136-145, 2008.
[62] K. Stephenson and M. Zelen, "Rethinking centrality: methods and examples", *Social Network*, 11(1):1-37, 1989.
[63] P. Hage, and F. Harary, "Eccentricity and centrality in networks", *Social Networks*, 17(1):57-63, 1995.
[64] J. D. Noh, and H. Rieger, "Random walks on complex networks", *Physical Review Letter*, 92:11-19, 2004.
[65] T-D. Tran and Y-K. Kwon, "Hierarchical closeness efficiently predicts disease genes in a directed signaling network", *Computational Biology and Chemistry*, 53:191-197, 2014.
[66] P. Bonacich, "Factoring and weighting approaches to status scores and clique identification", *Journal of Mathematical Sociology*, 2:113-120, 1972.
[67] W. Richards and A. Seary, "Eigen analysis of networks", *Journal of Social Structure*, 1, 2000.
[68] L. Katz, "A new status index derived from sociometric analysis", *Psychometrika*, 18(1):39-43, 1953.
[69] L. Page, S. Brin, R. Motwani and T. Winograd, "The PageRank citation ranking: bringing order to the web", *Technical Report*, Stanford InfoLab. 1999.
[70] E. Estrada and J. A. Rodríguez-Velázquez, "Subgraph centrality in complex networks", *Physical Review E*, 71(5), 2005.
[71] J. A. Rodríguez, E. Estrada and A. Gutiérrez, "Functional centrality in graphs", *Linear and Multilinear Algebra*, 55(3):292-302, 2007.
[72] D. J. Watts and S. H. Strogatz, "Collective dynamics of 'small-world' networks", *Nature*, 393(6684):440-442, 1998.
[73] A-L. Barabási and R. Albert, "Emergence of scaling in random networks", *Science*, 286:509-512, 1999.
[74] P. Erdős and A. Rényi, "On random graphs I", *Publicationes Mathematica*, 6:290-297, 1959.
[75] J. Leskovec, D. Chakrabarti, J. Kleinberg, C. Faloutsos and Z. Ghahramani, "Kronecker graphs: an approach to modeling networks", *Journal of Machine Learning Research*, 11:985-1042, 2010.
[76] F. Chung and L. Lu, "Connected components in random graphs with given degree sequences", *Annals of Combinatorics*, 6:125, 2002.
[77] F. Chung, and L. Lu, "The average distances in random graphs with given expected degrees", *Proc. of the National Academy of Sciences*, 99:15879, 2002.
[78] D A. Bader, S. Kintali, K. Madduri and M. Mihail, "Approximating betweenness centrality", *Proc. Of the 5th International Conference on Algorithms and Models for the Web-Graph*, 124-137, 2007.
[79] D. Eppstein and J. Wang, "Approximating centrality", *Journal of Graph Algorithms and Applications*, 8(1):39-45, 2004.
[80] E. Cohen, D. Delling, T. Pajor and R. F. Werneck. "Computing classic closeness centrality, at scale". *ACM Conference on Online Social Networks*. 2014.
[81] U. Brandes and C. Pich, "Centrality estimation in large networks", *International Journal of Bifurcation and Chaos*, 17(7):1-30, 2007.
[82] J. Leskovec, J. Kleinberg and C. Faloutsos, "Graphs over time: densification laws, shrinking diameters and possible explanations", *Proc. of the ACM SIGKDD International Conference on Knowledge Discovery and Data Mining*, 2005.

[83] J. Leskovec, J. Kleinberg and C. Faloutsos, "Graph evolution: densification and shrinking diameters", *ACM Transactions on Knowledge Discovery from Data*, 1(1), 2007.
[84] J. Yang and J. Leskovec, "Defining and evaluating network communities based on ground-truth", *Proc. of the IEEE International Conference on Data Mining*, 2012.
[85] E. Cho, S. A. Myers and J. Leskovec, "Friendship and mobility: user movement in location-based social networks", *Proc. of the ACM SIGKDD International Conference on Knowledge Discovery and Data Mining*, 2011.
[86] J. Leskovec, K. Lang, A. Dasgupta and M. Mahoney, "Community structure in large networks: natural cluster sizes and the absence of large well defined clusters", *Internet Mathematics*, 6(1):29-123, 2009.
[87] B. Klimmt and Y. Yang, "Introducing the Enron corpus", *Proc. of the Council of European Aerospace Societies*, 2004.
[88] R. Zafarani and H. Liu, "Social computing data repository at ASU", Arizona State University, School of Computing, informatics and Decision Systems Engineering, 2009.
[89] M. Fire, L. Tenenboim-Chekina, R. Puzis, O. Lesser, L. Rokach and Y. Elovici, "Link prediction in social networks using computationally efficient topological features", *Proc. of the IEEE 3$^{rd}$ International Conference on Social Computing*, 2011.
[90] M. Fire, L. Tenenboim-Chekina, R. Puzis, O. Lesser, L. Rokach and Y. Elovici, "Computationally efficient link prediction in a variety of social networks", *ACM Transactions on Intelligent Systems and Technology*, 2013.
[91] Network Dataset, KONECT, at http://konect.uni-koblenz.de/, 2016.
[92] L. Šubelj and M. Bajec, "Robust network community detection using balanced propagation", *The European Physical Journal J. B.*, 81(3):353-362, 2011.
[93] M. Boguñá, R. Pastor-Satorras, A. Díaz-Guilera and A. Arenas, "Models of social networks based on social distance attachment", *Physical Review E*, 70(5), 2004.
[94] A. Kumar, K. G. Mehrotra and C. K. Mohan, "Neural networks for fast estimation of social network centrality measures", Proc. of the 5$^{th}$ International Conference on Fuzzy and Neuro Computing (FANCCO), 175-184, 2015.
[95] S. Boccaletti, V. Latora, Y. Moreno, M. Chavez, and D-U. Hwang, "Complex networks: Structure and dynamics", *Physics Reports*, 424:175-308, 2006.
[96] M. Mitzenmacher, "A brief history of generative models for power law and lognormal distributions", *Internet Mathematics*, 1(2):226-251, 2003.
[97] M. T. Hagan, H. B. Demuth and M. H. Beale, *Neural Network Design*, Boston, MA, PWS Publishing, 1996.
[98] R. Battiti, "First and second order methods for learning: Between steepest descent and Newton's method", *Neural Computation*, 4(2):141-166, 1992.
[99] J.E. Dennis and R. B. Schnabel, *Numerical Methods for Unconstrained Optimization and Nonlinear Equations*, Englewood Cliffs, NJ, Prentice-Hall, 1983.
[100] P. E. Gill, W. Murray and M. H. Wright, *Practical Optimization*, Bingley, UK, Emerald Publishing, 1981.
[101] L. E. Scales, *Introduction to Non-Linear Optimization*, New York, Springer-Verlag, 1985.
[102] C. Charalambous, "Conjugate gradient algorithm for efficient training of artificial neural networks", *IEEE Proc. G – Circuits, Devices and Systems*, 139(3):301-310, 1992.
[103] M. J. D. Powell, "Restart procedures for the conjugate gradient method", *Mathematical Programming*, 12:241-254, 1977.
[104] M. F. Møller, "A scaled conjugate gradient algorithm for fast supervised learning", "*Neural Networks*", 6:525-533, 1993.
[105] M. Riedmiller and H. Braun, "A direct adaptive method for faster backpropagation learning: The RPROP algorithm", "*Proc. of the IEEE International Conference on Neural Networks*", 586-591, 1993.
[106] D. J. C. MacKay, "Bayesian Interpolation", *Neural Computation*, MIT Press Journal, 4(3):415-447, 1992.
[107] M. T. Hagan and M. Menhaj, "Training feed-forward networks with the Marquardt algorithm", *IEEE Transactions on Neural Networks*, 5(6):989-993, 1994.
[108] Y. LeCun, Y. Bengio and G.E. Hinton. Deep Learning. *Nature* 521:14539, May 2015.
[109] Y. Bengio, A. Courville and P. Vincent. Representation Learning: A Review and New Perspectives. IEEE Transactions on Pattern Analysis and Machine Intelligence 35(8):1798-1898, August 2013.